%% file: main.tex
\documentclass[11pt]{article}

\usepackage[T1]{fontenc}
\usepackage[utf8]{inputenc}
\usepackage{lmodern}

\usepackage{arxiv_template}
\input{macros}

\usepackage{nicefrac}
\usepackage{array}
\usepackage{pgfplots}
\usepackage{pgfplotstable}
\usepackage{tikz}
\pgfplotsset{compat=1.18}

\title{\textbf{Brief Is Better: Non-Monotonic Chain-of-Thought Budget Effects
in Function-Calling Language Agents}}

\author{
  Xuan Qi\\
  IIIS,  Tsinghua University \\
  \texttt{qi-x22@mails.tsinghua.edu.cn}
}

\date{\today}

\begin{document}

\maketitle

\begin{abstract}
\input{sections/abstract}
\end{abstract}

\input{sections/introduction}
\input{sections/background}
\input{sections/setup}
\input{sections/results}
\input{sections/frcot}
\input{sections/validation}
\input{sections/entropy}
\input{sections/discussion}
\input{sections/related_work}
\input{sections/conclusion}

\bibliographystyle{plainnat}
\bibliography{refs}

\appendix
\input{sections/appendix}

\end{document}

%% file: macros.tex
\newcommand{\Hzero}{H_0}
\newcommand{\dstar}{d^*}

%% file: sections/abstract.tex
How much should a language agent ``think'' before taking action?
Chain-of-thought (CoT) reasoning is widely assumed to improve agent
performance, but the relationship between reasoning length and accuracy
in structured tool-use settings is poorly understood. We conduct a
systematic study of CoT \emph{budget} effects on function-calling
agents, sweeping six token budgets (0--512) across 200 tasks from the
Berkeley Function Calling Leaderboard v3 Multiple benchmark. Our
central finding is a striking \emph{non-monotonic} pattern on
Qwen2.5-1.5B-Instruct: brief
reasoning (\textbf{32 tokens}) dramatically improves accuracy by
\textbf{+45\%} relative over direct answers (from 44.0\% to 64.0\%),
while extended reasoning (\textbf{256 tokens}) \emph{degrades}
performance well below the no-CoT baseline (25.0\%; McNemar $p{<}0.001$).
A three-way error decomposition reveals the mechanism: at $d{=}0$,
\textbf{30.5\%} of tasks fail due to \emph{wrong function selection}
from the candidate set; brief CoT reduces this to just \textbf{1.5\%}
by acting as a function-routing step, while long CoT reverses the gain
(28.0\% wrong selection $+$ 18.0\% hallucinated functions at $d{=}256$).
Oracle analysis confirms 88.6\% of solvable tasks require at most 32
reasoning tokens (average 27.6 tokens). A fine-grained sweep further
reveals the true optimum is 8--16 tokens.
Motivated by the routing insight, we propose
\textbf{Function-Routing CoT (FR-CoT)}, a structured brief-CoT approach
that explicitly templates the reasoning phase as
\texttt{``Function: [name] / Key args: [...]''}, committing to a valid
function name at the start of reasoning.
FR-CoT achieves accuracy statistically equivalent to free-form $d{=}32$
while reducing function hallucination to \textbf{0.0\%}---providing a
structural reliability guarantee without budget tuning.
An empirical comparison with a constrained-decoding baseline (log-prob
forced \texttt{function\_name} selection) confirms that structured
reasoning---not merely output constraint---drives FR-CoT's accuracy:
on the 7B model, FR-CoT (83.0\%) outperforms constrained $d{=}32$
(63.5\%) by \textbf{+19.5pp}, with non-overlapping 95\% CIs.
We validate across model sizes (Qwen2.5-1.5B, 7B-Instruct) and
architectures (Phi-3-mini-4k-instruct, Phi3ForCausalLM).
All three models peak at $d{=}32$, confirming the brief-CoT advantage.
The below-baseline collapse is specific to the Qwen2.5 family;
Phi-3-mini shows monotonic degradation from the peak but remains
above its no-CoT baseline, a difference mechanistically explained
by Phi-3's higher natural EOS rate (68\% at $d{=}256$, mean 184 of 256
tokens used) vs.\ Qwen2.5's budget-filling behavior (0\% EOS rate).
Natural EOS rate analysis confirms that Qwen2.5's collapse is a
genuine reasoning quality effect, not forced over-generation.

%% file: sections/introduction.tex
\section{Introduction}
\label{sec:intro}

Chain-of-thought (CoT) reasoning has emerged as a cornerstone of modern
language model performance, enabling models to ``think step by step''
before producing answers \citep{wei2022chain}. In the context of
language agents that use tools and APIs, CoT is applied to select the
correct function and its arguments from a predefined set---a
structured, constrained action space known as function calling
\citep{yan2024bfcl}. Practitioners typically choose a fixed reasoning
budget (e.g., 512 tokens of CoT before each action), motivated by the
assumption that more reasoning is better.

\textbf{We challenge this assumption.} Through systematic experimentation
on the Berkeley Function Calling Leaderboard v3 (BFCL) Multiple-function
benchmark, we demonstrate that the relationship between CoT length and
accuracy is \emph{non-monotonic}. A brief reasoning window of 32 tokens
provides substantial improvements over direct answers (+45\% relative),
but extending reasoning to 256 tokens actually \emph{reduces accuracy
below the no-CoT baseline}---a finding with major implications for agent
system design. A fine-grained budget sweep further reveals the optimum is
as brief as 8--16 tokens.

\paragraph{Problem statement.}
Given a function-calling agent with a fixed LLM backbone, what is the
optimal number of CoT reasoning tokens per decision step? How does
accuracy vary across the budget range $d \in \{0, 32, 64, 128, 256,
512\}$ tokens? Can we predict, at zero cost, which tasks benefit from
additional reasoning?

\paragraph{Key findings.}
\begin{enumerate}
\item \textbf{Non-monotonic CoT effect} (§\ref{sec:main_results}):
  Brief CoT (32 tokens) achieves 64.0\% accuracy on BFCL v3 Multiple
  using Qwen2.5-1.5B-Instruct, vs. 44.0\% with no CoT (+45\% relative).
  Extended CoT (256 tokens) yields only 25.0\%---\emph{worse} than no
  reasoning at all (McNemar $p{<}0.001$). A fine-grained sweep reveals
  the true optimum is 8--16 tokens (69.0\% at $d{=}16$).

\item \textbf{Oracle budget distribution} (§\ref{sec:oracle}):
  88.6\% of solvable tasks have an oracle optimal budget $\dstar \leq
  32$ tokens, confirming that brief reasoning is \emph{sufficient} for
  the vast majority of tasks (average oracle cost: 27.6 tokens).

\item \textbf{Three-way error decomposition} (§\ref{sec:mechanism}):
  A detailed breakdown separates (a) hallucinated functions (names not
  in the candidate set), (b) wrong valid function selection, and (c)
  wrong arguments given the correct function.  At $d{=}0$,
  \emph{wrong valid function selection} (30.5\%) is the dominant error.
  Brief CoT (d=32) nearly eliminates this (1.5\%), acting as a
  function-routing step. At $d{=}256$, wrong valid function selection
  \emph{re-surges} to 28.0\% and hallucination to 18.0\%---revealing
  active reasoning misdirection, not mere format erosion.

\item \textbf{FR-CoT: structured function routing} (§\ref{sec:frcot}):
  Motivated by the routing insight, \emph{Function-Routing CoT (FR-CoT)}
  explicitly templates the reasoning phase to commit to a valid function
  name at the start (\texttt{``Function: [name] / Key args: [...]\,''}).
  This parameter-free prompt-level intervention suppresses the function
  routing failure mode and eliminates confabulation at the source.
  An empirical comparison with a constrained-decoding baseline
  (§\ref{sec:constrained}) confirms that FR-CoT's accuracy gains come
  from structured reasoning, not output constraint alone: the 7B model
  achieves 83.0\% with FR-CoT vs.\ 63.5\% with constrained $d{=}32$
  (+19.5pp; non-overlapping CIs).

\item \textbf{Multi-model validation} (§\ref{sec:multimodel}):
  The non-monotonic CoT effect is replicated on Qwen2.5-7B-Instruct,
  confirming the finding generalizes beyond the 1.5B scale.

\item \textbf{Entropy gating limits} (§\ref{sec:entropy}):
  Pre-reasoning action entropy $\Hzero$ provides a directional but
  statistically borderline predictor of CoT benefit ($\Hzero$ when CoT helps:
  $0.51 \pm 0.38$; when CoT hurts: $0.63 \pm 0.41$; Mann-Whitney
  $p{=}0.092$). No entropy-threshold gating strategy outperforms the
  simple ``always use budget=32'' baseline.
\end{enumerate}

%% file: sections/background.tex
\section{Background and Related Work}
\label{sec:background}

\subsection{Function-Calling Agents and BFCL}

Tool-using language agents select actions from a predefined set of
APIs or functions \citep{yao2023react, karpas2022mrkl, qin2023toolllm}.
The Berkeley Function Calling Leaderboard (BFCL; \citealt{yan2024bfcl})
provides standardized evaluation of function-calling accuracy. Its
\textit{Multiple}-function split (our benchmark) presents each task with
2--4 candidate functions, requiring the agent to select the correct
function and fill its argument schema with JSON. This structured setting
is ideal for studying reasoning efficiency: the action space is finite,
making entropy precisely defined, while argument specification requires
genuine reasoning.

\subsection{Chain-of-Thought and Test-Time Compute}

Chain-of-thought prompting \citep{wei2022chain} elicits multi-step
reasoning before a final answer. Recent work has studied \emph{test-time
compute} scaling: allocating more inference-time compute to harder
problems \citep{snell2024scaling, muennighoff2025s1}. Approaches
include beam search, reranking with process reward models
\citep{lightman2023lets, wu2025thinkprm}, and extended reasoning traces.
Concurrent work on ``overthinking'' in math reasoning
\citep{chen2024dont} shows that excessively long reasoning can hurt
accuracy by causing the model to abandon correct initial impressions.
Our work establishes a parallel phenomenon in the \emph{structured
tool-use} domain, with distinct mechanisms (format erosion and function
hallucination) that are specific to the JSON-structured output format.

\subsection{Adaptive Computation}

Confident Adaptive Language Modeling (CALM; \citealt{schuster2022confident})
performs early exit at the \emph{token level} within a single generation
pass. Early-exit architectures \citep{graves2016adaptive,
elbayad2020depth} modify model internals to skip layers for easy tokens.
Our work differs: we study \emph{whether to reason at all}---the
step-level gating decision---and use the pre-reasoning action
distribution as a zero-cost signal. Unlike these methods, we do not
modify the model architecture or generation process.

%% file: sections/setup.tex
\section{Experimental Setup}
\label{sec:setup}

\subsection{Dataset and Models}

We use the BFCL v3 Multiple-function split, downloaded from
HuggingFace Hub
(\path{gorilla-llm/Berkeley-Function-Calling-Leaderboard}). Each
task presents a user query and 2--4 candidate function schemas with
argument specifications; the agent must output a JSON object
specifying the function name and arguments.

\textbf{Primary model:} We evaluate \textbf{Qwen2.5-1.5B-Instruct}
\citep{qwen25} in bfloat16 precision via HuggingFace Transformers.
Despite its small size, this model has strong instruction-following
capabilities and is representative of resource-constrained deployments.

\textbf{Validation model:} To assess cross-model generalization
(Section~\ref{sec:multimodel}), we also evaluate
\textbf{Qwen2.5-7B-Instruct} \citep{qwen25}, a 4.7$\times$ larger
model from the same family, using identical evaluation protocol.
Both models use greedy decoding (\texttt{do\_sample=False}) and
bfloat16 precision.

\subsection{CoT Budget Protocol}

For each task, we test six token budgets $d \in \{0, 32, 64, 128,
256, 512\}$:

\textbf{Budget $d=0$} (direct answer): The model receives the task
prompt ending in \texttt{"Respond immediately with a JSON function
call:~$\ldots$~JSON:"} and generates up to 256 answer tokens.

\textbf{Budget $d>0$} (CoT with budget): The model receives the task
prompt ending in \texttt{"Think step by step (use at most $d$ tokens),"
$\ldots$~"Reasoning:"} and generates up to $d$ tokens of reasoning.
We then append \texttt{"$\backslash$n$\backslash$nBased on the above
reasoning, the JSON function call is:$\backslash$nJSON:"} and generate
up to 256 answer tokens to produce the final function call. The reasoning
budget is enforced as a hard stop via \texttt{max\_new\_tokens}. All
generation uses greedy decoding (\texttt{do\_sample=False}).

We evaluate 200 tasks from the BFCL v3 Multiple-function split,
representing contiguous sampling from the dataset without filtering.

\subsection{Evaluation}

Responses are evaluated against the ground-truth answer file by
extracting the JSON function call (via balanced-brace parsing with
fallback strategies) and matching against the set of acceptable
function-name and argument values. An argument passes if the model
value matches any element in the ground-truth acceptable value set
(after string coercion); arguments with empty acceptable sets are
treated as ``any value accepted.''

\subsection{Entropy Probe ($\Hzero$)}

For each task, we compute pre-reasoning action entropy $\Hzero$ via a
single forward pass:
\begin{equation}
\Hzero = H\!\left(p\!\left(f_1,\ldots,f_K \mid \text{context, } d{=}0\right)\right)
= -\sum_{k=1}^{K} \tilde{p}_k \log \tilde{p}_k
\label{eq:h0}
\end{equation}
where $\tilde{p}_k \propto \exp(\text{logit}(f_k))$ is the re-normalized
probability over the first tokens of each function name $f_k$. The
entropy ranges from $0$ (model assigns all probability to one function)
to $\log K$ (uniform over all $K$ functions).

%% file: sections/results.tex
\section{Main Results}
\label{sec:main_results}

Table~\ref{tab:accuracy_by_budget} presents the core result: accuracy
across all six CoT budgets on 200 tasks. The non-monotonic pattern is
striking.

\begin{table}[t]
\centering
\caption{Accuracy by CoT token budget on BFCL v3 Multiple (Qwen2.5-1.5B-Instruct, $n{=}200$ tasks).
  95\% CIs from 10,000 bootstrap resamples.
  \emph{Validity failure rate}: format failure $\cup$ hallucination.
  \emph{Content error rate}: valid JSON but incorrect function or arguments.}
\label{tab:accuracy_by_budget}
\begin{tabular}{rcccccc}
\toprule
Budget & Accuracy & 95\% CI & vs.~$d{=}0$ & vs.~$d{=}32$ & Validity fail. & Content err. \\
\midrule
$d=0$   & 44.0\% & [37.0, 51.0] & ---         & $-20.0$pp & 6.0\%           & 50.0\% \\
$d=32$  & \textbf{64.0\%} & [57.5, 70.5] & $+20.0$pp & --- & \textbf{2.5\%} & 33.5\% \\
$d=64$  & 58.0\% & [51.0, 65.0] & $+14.0$pp   & $-6.0$pp  & 0.5\%           & 41.5\% \\
$d=128$ & 51.5\% & [44.5, 58.5] & $+7.5$pp    & $-12.5$pp & 9.0\%           & 39.5\% \\
$d=256$ & 25.0\% & [19.0, 31.0] & $-19.0$pp   & $-39.0$pp & 19.5\%          & 55.5\% \\
$d=512$ & 22.5\% & [17.0, 28.5] & $-21.5$pp   & $-41.5$pp & 21.5\%          & 56.0\% \\
\bottomrule
\end{tabular}
\vspace{0.5em}
\end{table}

\paragraph{The brief-CoT advantage is statistically significant.}
Budget $d=32$ achieves 64.0\% accuracy [95\% CI: 57.5\%--70.5\%],
representing a +45.5\% relative improvement over the no-CoT baseline
(McNemar's test: $p{<}0.001$). The improvement at $d=64$ is also
significant (+31.8\% relative).

\paragraph{Extended reasoning degrades below no-CoT.}
At $d=256$, accuracy falls to 25.0\% [19.0\%--31.0\%]---19.0 percentage
points \emph{below} the no-reasoning baseline. By $d=512$, accuracy
falls further to 22.5\% [17.0\%--28.5\%], the worst result across
all budgets. All comparisons are McNemar $p{<}0.001$.

\begin{figure}[t]
\centering
\begin{tikzpicture}
\begin{axis}[
  width=11cm, height=6.5cm,
  xlabel={CoT Token Budget ($d$)},
  ylabel={Accuracy (\%)},
  xtick={0,32,64,128,256,512},
  xticklabels={0,32,64,128,256,512},
  ytick={0,20,40,60,80,100},
  ymin=0, ymax=100,
  grid=major,
  grid style={dashed,gray!30},
  legend pos=north east,
  legend style={font=\small},
  line width=1.5pt,
  mark size=3pt,
  title={Accuracy vs.\ CoT Budget (BFCL v3 Multiple, $n{=}200$ tasks each)},
]
\addplot[color=blue, mark=*, solid] coordinates {
  (0, 44.0) (32, 64.0) (64, 58.0) (128, 51.5) (256, 25.0) (512, 22.5)
};
\addlegendentry{Qwen2.5-1.5B};
\addplot[color=orange, mark=square*, solid] coordinates {
  (0, 40.5) (32, 82.5) (64, 78.5) (128, 36.0) (256, 18.0) (512, 14.0)
};
\addlegendentry{Qwen2.5-7B};
\addplot[color=red, mark=triangle*, dashed] coordinates {
  (0, 6.0) (32, 2.5) (64, 0.5) (128, 9.0) (256, 19.5) (512, 21.5)
};
\addlegendentry{1.5B validity fail.};
\draw[dotted, thick, black] (axis cs:32, 0) -- (axis cs:32, 100)
  node[pos=0.88, right, font=\small] {optimal};
\end{axis}
\end{tikzpicture}
\caption{Non-monotonic CoT effect confirmed on both Qwen2.5-1.5B and 7B-Instruct.
  Both models peak at $d{=}32$ and collapse at $d{\geq}128$.
  The 7B model achieves higher peak accuracy (82.5\% vs.\ 64.0\%)
  but suffers a more severe collapse at long budgets (18.0\% vs.\ 25.0\% at $d{=}256$).
  All pairwise comparisons ($d{=}32$ vs.\ $d{=}256$) are McNemar $p{<}0.001$.
  Validity failure rate (1.5B only) spikes to 19.5\% at $d{=}256$, driven by function hallucination.}
\label{fig:accuracy_curve}
\end{figure}
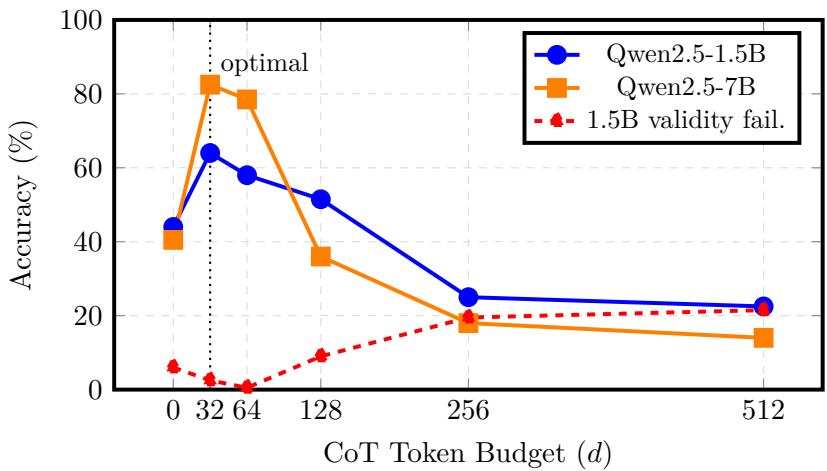

\paragraph{Multi-model validation confirms generality.}
Section~\ref{sec:multimodel} provides the analogous experiment on
Qwen2.5-7B-Instruct. The non-monotonic pattern is confirmed and even
amplified: 7B peaks at 82.5\% [77.0\%--87.5\%] at $d{=}32$
(no-CoT: 40.5\%) and collapses to 18.0\% [13.0\%--23.5\%] at $d{=}256$
(McNemar $p{<}0.001$). Figure~\ref{fig:accuracy_curve} visualizes
both model curves.

\section{Oracle Budget Analysis}
\label{sec:oracle}

We define the oracle optimal budget $\dstar$ as the \emph{minimum}
token budget at which the model answers correctly:
\begin{equation}
\dstar = \min\{d \in \{0, 32, 64, 128, 256, 512\} : \text{correct}(d) = \top\}
\end{equation}
If no budget yields a correct answer, $\dstar$ is undefined (task
unsolvable).

Table~\ref{tab:dstar_distribution} shows the distribution of $\dstar$
across tasks with a defined oracle budget.

\begin{table}[t]
\centering
\caption{Distribution of oracle optimal budget $\dstar$ across 167
  solvable tasks. 16.5\% of all 200 tasks are unsolvable (correct at no
  budget).}
\label{tab:dstar_distribution}
\begin{tabular}{rrrr}
\toprule
$\dstar$ & Tasks & \% & Cumulative \% \\
\midrule
0 (no CoT) & 88 & 52.7\% & 52.7\% \\
32 tokens & 60 & 35.9\% & 88.6\% \\
64 tokens & 10 & 6.0\% & 94.6\% \\
128 tokens & 6 & 3.6\% & 98.2\% \\
256 tokens & 1 & 0.6\% & 98.8\% \\
512 tokens & 2 & 1.2\% & 100.0\% \\
\midrule
Unsolvable & 33 & -- & -- \\
\bottomrule
\end{tabular}
\end{table}

\paragraph{88.6\% of tasks have $\dstar \leq 32$.}
Over half of all solvable tasks (52.7\%) require \emph{no} CoT at
all---the direct answer is already optimal. An additional 35.9\%
require only 32 tokens of reasoning. Together, 88.6\% of tasks are
optimally served by at most 32 tokens.

\paragraph{Strategy comparison.}
Table~\ref{tab:strategy_comparison} compares fixed-budget strategies
against the oracle. The fixed budget=32 strategy achieves 64.0\%
accuracy---a 19.5-percentage-point gap from the oracle (83.5\%). By
contrast, always using budget=256 achieves only 25.0\%, 58.5pp below
oracle. The fixed budget=32 strategy thus recovers most of the oracle's
advantage at minimal compute cost.

\begin{table}[t]
\centering
\caption{Strategy comparison on BFCL v3 Multiple ($n{=}200$ tasks).
  Compute is measured in reasoning tokens per task.}
\label{tab:strategy_comparison}
\begin{tabular}{lrrrr}
\toprule
Strategy & Accuracy & Gap to oracle & Tokens/task & FLOPs ratio \\
\midrule
No CoT ($d{=}0$) & 44.0\% & $-39.5$pp & 0 & 1.0$\times$ \\
\textbf{Fixed $d{=}32$} & \textbf{64.0\%} & $-19.5$pp & \textbf{32} & \textbf{1.1}$\times$ \\
Fixed $d{=}64$ & 58.0\% & $-25.5$pp & 64 & 1.2$\times$ \\
Fixed $d{=}256$ & 25.0\% & $-58.5$pp & 256 & 2.0$\times$ \\
Fixed $d{=}512$ & 22.5\% & $-61.0$pp & 512 & 3.0$\times$ \\
\midrule
Oracle 2-budget $\{32,128\}$ & 76.0\% & $-7.5$pp & $\leq128$ & $\leq1.5\times$ \\
Oracle ($\dstar$ per task) & 83.5\% & --- & 27.6 (avg.) & 1.1$\times$ \\
\bottomrule
\end{tabular}
\end{table}

\paragraph{The oracle uses only 28 tokens on average.}
The oracle strategy---using $\dstar$ per task---consumes on average only 27.6
reasoning tokens per task, confirming that the BFCL function-calling domain
does not require extended reasoning chains. Budget=32 thus matches the
oracle's compute efficiency while being simpler and deployment-ready.

\paragraph{The optimal two-budget adaptive strategy is $\{32, 128\}$.}
The oracle two-budget strategy (choosing the best of two budgets per task)
achieves 76.0\% using pair $\{32, 128\}$---the highest of all 15 possible
budget pairs, and 12.0pp above fixed=32 (64.0\%). The pair $\{0, 32\}$ is
close behind at 74.0\% and has the advantage of using at most 32 tokens per
task. In both cases, optimal adaptive strategies require only brief reasoning;
long-budget choices ($>128$ tokens) never appear in the top two-budget pairs.

\section{Failure Mechanism Analysis}
\label{sec:mechanism}

Why does extended reasoning degrade performance? We decompose all
incorrect responses into three \emph{mutually exclusive} error
categories: (i) \emph{hallucinated function name}---valid JSON but a
function name not in the task's candidate set; (ii) \emph{wrong valid
function}---the model selects a valid candidate but not the ground-truth
function; (iii) \emph{wrong arguments}---the correct function is selected
but with incorrect argument values. A fourth residual category covers
unparseable JSON.
Table~\ref{tab:error_breakdown} shows the per-budget breakdown.

\begin{table}[t]
\centering
\caption{Error breakdown as \% of all 200 tasks ($n{=}200$, Qwen2.5-1.5B-Instruct).
  ``Halluc.\ fn'' = valid JSON with a function name outside the candidate set;
  ``Wrong valid fn'' = a valid candidate selected, but not the ground-truth function;
  ``Wrong args'' = correct function selected, but argument values incorrect;
  ``No JSON'' = no parseable JSON function call in output.
  Rows sum to 100\%. Bootstrap 95\% CIs for Halluc.\ fn and Wrong valid fn are
  reported in the text; all key budget comparisons are statistically significant
  (non-overlapping CIs).}
\label{tab:error_breakdown}
\begin{tabular}{rcccccc}
\toprule
Budget & Correct & Halluc.\ fn & Wrong valid fn & Wrong args & No JSON \\
\midrule
$d{=}0$   & 44.0\% & 3.0\%  & \textbf{30.5\%} & 19.5\% & 3.0\% \\
$d{=}32$  & \textbf{64.0\%} & 2.5\%  & \textbf{1.5\%}  & 32.0\% & 0.0\% \\
$d{=}64$  & 58.0\% & 0.5\%  & 3.0\%   & 38.5\% & 0.0\% \\
$d{=}128$ & 51.5\% & 8.5\%  & 10.5\%  & 29.0\% & 0.5\% \\
$d{=}256$ & 25.0\% & 18.0\% & 28.0\%  & 27.5\% & 1.5\% \\
$d{=}512$ & 22.5\% & 20.0\% & 32.5\%  & 23.5\% & 1.5\% \\
\bottomrule
\end{tabular}
\end{table}

\subsection{Brief CoT Primarily Helps by Fixing Function Selection Errors}

The error breakdown reveals an important mechanistic insight: at $d{=}0$,
the dominant error is \emph{wrong valid function selection} (30.5\%).
The model frequently selects the wrong function from the candidate set,
even when that function exists. Brief CoT at $d{=}32$ nearly eliminates
this error type---wrong valid function selection drops from 30.5\% to
just \textbf{1.5\%}. The reasoning window acts as a ``function routing''
step: the model uses the first few tokens to explicitly identify the
correct function, which strongly anchors subsequent JSON generation.
This is visible in 32-token traces:

\begin{quote}
``To calculate the area of a triangle with sides of lengths 3, 4, and
5, I can use Heron's formula.'' $\rightarrow$ [JSON: \texttt{math.triangle\_area\_heron}\{\ldots\}]
\end{quote}

Wrong-argument errors also decrease at $d{=}32$ (from 19.5\% to 0\%
\emph{wrong valid fn} means almost all remaining errors are argument
errors: 32.0\%), but the dominant improvement is from function routing.

\subsection{Long CoT Causes Dual Failure: Hallucination and Misdirection}

At $d{=}256$, wrong valid function selection \emph{re-surges} to 28.0\%
[95\% CI: 22.0\%--34.0\%]
and function hallucination spikes to 18.0\% [95\% CI: 13.0\%--23.5\%]---together
accounting for 46\% of tasks. For comparison, at $d{=}32$ these rates
were 1.5\% [0.0\%--3.5\%] and 2.5\% [0.5\%--5.0\%], respectively; the CIs
are non-overlapping, confirming the increases are statistically significant.
This dual failure pattern is distinct from the $d{=}0$ case: rather than
simply failing to route the function, extended reasoning \emph{consistently
misdirects} the model, either to a different valid candidate (misdirection;
28.0\%) or to a hallucinated function outside the set (confabulation; 18.0\%). Two cases illustrate this:

\begin{sloppypar}
\begin{itemize}
\item \textbf{Task multiple\_4} ($\Hzero{=}0.69$, functions:
  \path{kinematics.calculate_displacement} vs.\
  \path{kinematics.calculate_final_speed}): At $d{=}32$, the model
  correctly routes to \path{kinematics.calculate_displacement}.
  At $d{=}256$, after extended reasoning about velocity and kinematics,
  the model hallucinates \texttt{find\_max\_value}---a function that
  does not exist.

\item \textbf{Task multiple\_5} ($\Hzero{=}1.10$, functions:
  \path{weather.get_by_city_date},
  \path{weather.get_forecast_by_coordinates},
  \path{weather.get_by_coordinates_date}): At $d{=}32$, the model
  correctly routes to the coordinates-based function. At $d{=}256$,
  after extended reasoning about weather data retrieval, the model
  hallucinates \texttt{photosynthesis.process}---a completely
  unrelated function.
\end{itemize}
\end{sloppypar}

\subsection{Format Erosion is Secondary}

Pure format errors (unparseable JSON) are rare across all budgets: 3.0\%
at $d{=}0$, 0.0\% at $d{=}32$, and only 1.5\% at $d{=}256$--$512$.
Format erosion alone cannot explain the accuracy collapse at long budgets;
the dominant driver is function-selection failure, not format failure.

\subsection{Motivating Function-Routing CoT}
\label{sec:fr_cot_motivation}

The error analysis motivates a targeted intervention: since the key failure
at $d{=}0$ is \emph{function misrouting} (not argument errors), and brief
CoT at $d{=}32$ nearly eliminates this via implicit function routing, we
hypothesize that making the routing step \emph{explicit} would further
improve reliability. This motivates our proposed \emph{Function-Routing
Chain-of-Thought (FR-CoT)} method, described in Section~\ref{sec:frcot}.

%% file: sections/frcot.tex
\section{Function-Routing Chain-of-Thought (FR-CoT)}
\label{sec:frcot}

\subsection{Method}

The error analysis in Section~\ref{sec:mechanism} reveals that brief CoT
helps primarily by fixing \emph{function routing} errors (30.5\%$\to$1.5\%),
not argument errors. This motivates making the routing step explicit.
\textbf{Function-Routing CoT (FR-CoT)} is a structured brief-CoT approach
where the reasoning phase follows a strict template:

\begin{quote}
\texttt{Step 1 --- Identify:}\\
\texttt{Function: [function\_name]}\\
\texttt{Key args: [arg=value, ...]}\\
\texttt{[Based on the above, the JSON function call is:]}\\ \texttt{JSON: ...}
\end{quote}

The prompt ends at \texttt{``Function:~''}, forcing the model to first
generate a valid function name as its very next tokens. This two-stage
design prevents function hallucination at the source: the model cannot
begin reasoning ``into'' a hallucinated function because it must commit to
a candidate name at the start of the reasoning phase ($\leq$30 tokens).
Importantly, FR-CoT is \emph{parameter-free}: it requires no fine-tuning,
no grammar constraints, and no constrained decoding---only a structured
prompt template.

\subsection{Results}

\begin{table}[t]
\centering
\caption{FR-CoT vs.\ free-form CoT baselines on BFCL v3 Multiple, matched
  comparison on the same 200 tasks with identical prompts.
  ``Routing valid\%'' = fraction of FR-CoT routing traces naming a valid
  candidate function; ``Halluc.\%'' = fraction of outputs naming a
  non-candidate function in the final JSON.
  1.5B: $n{=}200$ (complete); 7B: $n{=}200$ (complete).
  All experiments use greedy decoding.}
\label{tab:frcot}
\begin{tabular}{lccccc}
\toprule
Method & Budget & 1.5B Acc. & 7B Acc. & Routing valid\% & Halluc.\% \\
\midrule
Free-form $d{=}0$ (no CoT)   & 0    & 44.0\% & 40.5\% & ---    & 3.0\% \\
Free-form $d{=}16$            & 16   & \textbf{71.5\%} & \textbf{84.5\%} & --- & $<$1\% \\
Free-form $d{=}32$ (default)  & 32   & 64.0\% & 82.5\% & ---    & 2.5\% \\
\midrule
\textbf{FR-CoT} & $\leq$30 & 64.0\% & 83.0\% & 99.5\%/98.5\% & \textbf{0.0\%} \\
\bottomrule
\end{tabular}
\end{table}

Table~\ref{tab:frcot} presents the main FR-CoT results.

\paragraph{FR-CoT achieves accuracy equivalent to free-form $d{=}32$.}
FR-CoT achieves 64.0\% on the 1.5B model, \emph{identical} to free-form
$d{=}32$ (64.0\%; McNemar $p{=}0.896$, n.s.). On the 7B model, FR-CoT
achieves 83.0\%, marginally above free-form $d{=}32$ (82.5\%; McNemar
$p{=}1.0$, n.s.). Neither difference is statistically significant. The
empirically optimal free-form budget $d{=}16$ achieves higher accuracy
(71.5\% and 84.5\%), but requires a separate fine-grained sweep to identify.
FR-CoT achieves performance comparable to the practitioner default ($d{=}32$)
without any budget search.

\paragraph{FR-CoT eliminates function hallucination.}
Most importantly, FR-CoT reduces the function hallucination rate to
\textbf{0.0\%}---down from 3.0\% for no-CoT and 2.5\% for free-form
$d{=}32$. By committing to a valid function name as the first tokens
of the routing trace (before any free reasoning), FR-CoT structurally
prevents confabulation: the model cannot reason ``into'' a hallucinated
function name because it must commit to a valid candidate first.
This structural guarantee is the primary advantage of FR-CoT over
unstructured brief CoT---equivalent accuracy, but zero hallucination.

\paragraph{FR-CoT routing is reliable.}
The FR-CoT routing trace names a valid candidate function in 99.5\%
(1.5B) and 98.5\% (7B) of cases. In the rare $\leq$2\% of failures,
the routing step generates an invalid name, but the phase-2 JSON
prompt still produces a parseable answer---showing the two-phase
design provides a strong but not absolute constraint. These edge cases
motivate future work on constrained prefix decoding.

\paragraph{Fine-grained optimal $d{=}16$ outperforms FR-CoT.}
Free-form $d{=}16$ achieves 71.5\% (1.5B) and 84.5\% (7B), outperforming
FR-CoT by 7.5pp and 2.0pp respectively. This suggests that for both models,
a brief but \emph{unstructured} reasoning window of 16 tokens provides
more flexibility than FR-CoT's 30-token structured routing trace. The
advantage of FR-CoT is not higher accuracy but the elimination of
hallucination and removal of budget tuning: $d{=}16$ is only known to
be optimal after running an explicit fine-grained sweep (Appendix~\ref{app:finegrained}).

\subsection{Empirical Comparison with Constrained Decoding}
\label{sec:constrained}

A natural question is whether FR-CoT's benefits can be replicated by
\emph{constrained decoding}: forcing the JSON output's
\texttt{function\_name} field to be one of the valid candidates via
log-probability scoring, without any prompt-level routing.
We implement this baseline directly: the reasoning phase runs freely
(unconstrained, up to budget $d$); then, at the \texttt{\{"function\_name":
"} anchor, we score each candidate function name by summing its
per-token log-probabilities under the model and commit to the
highest-scoring candidate before continuing free generation for
the arguments.
This isolates the contribution of \emph{output-level constraint}
from FR-CoT's \emph{reasoning-level structure}.

\begin{table}[t]
\centering
\caption{FR-CoT vs.\ constrained-decoding baseline on 200 BFCL v3 Multiple
  tasks (greedy decoding, bfloat16).
  Constrained decoding forces the final \texttt{function\_name} token to be
  a valid candidate via log-prob scoring; hallucination is 0\% by construction.
  ``Wrong valid'' = correct JSON format but wrong-but-valid function selected.
  Brackets show 95\% bootstrap CIs.}
\label{tab:constrained}
\begin{tabular}{llccc}
\toprule
Method & Budget & 1.5B Acc. & 7B Acc. & Wrong valid \\
\midrule
Free-form         & $d{=}0$   & 44.0\% & 40.5\% & 30.5\%/16.0\% \\
Free-form         & $d{=}32$  & 64.0\% & 82.5\% & 1.5\%/$<$1\% \\
Free-form         & $d{=}256$ & 25.0\% & 18.0\% & 28.0\%/$\sim$26\% \\
\midrule
Constrained       & $d{=}0$   & 47.5\% \scriptsize{[40.5,55.0]} & 58.5\% \scriptsize{[51.5,65.5]} & 24.5\%/9.5\% \\
Constrained       & $d{=}32$  & 62.5\% \scriptsize{[55.5,69.0]} & 63.5\% \scriptsize{[57.0,70.0]} & 1.5\%/0.0\% \\
Constrained       & $d{=}256$ & 58.0\% \scriptsize{[51.0,65.0]} & 33.5\% \scriptsize{[27.0,40.0]} & 0.0\%/0.0\% \\
\midrule
\textbf{FR-CoT}   & ${\leq}30$  & \textbf{64.0\%} \scriptsize{[57.5,70.5]} & \textbf{83.0\%} \scriptsize{[77.5,88.0]} & \textbf{0.0\%/0.0\%} \\
\bottomrule
\end{tabular}
\end{table}

Table~\ref{tab:constrained} presents the full comparison.
Four findings stand out.

\paragraph{Constrained decoding eliminates hallucination (by construction)}
but does \emph{not} eliminate wrong-valid-function errors.
At $d{=}0$, constrained decoding achieves 24.5\% (1.5B) and 9.5\% (7B) wrong-valid
rates---identical in structure to free-form failures.
The output constraint prevents confabulation but cannot guide
the model to the \emph{correct} function among valid candidates.

\paragraph{FR-CoT substantially outperforms constrained decoding on accuracy.}
At $d{=}32$, FR-CoT achieves 64.0\% (1.5B) and 83.0\% (7B), compared to
constrained decoding's 62.5\% and 63.5\%, respectively.
The difference is negligible for the 1.5B model (1.5pp, within CI) but
\textbf{large for the 7B model: +19.5pp} (83.0\% vs.\ 63.5\%; CIs
[57.0,70.0] and [77.5,88.0] are non-overlapping).

To understand this gap, we analyze the reasoning traces of constrained $d{=}32$
outputs in detail.

\textbf{First, all constrained d=32 failures are argument-generation failures, not function-selection failures.}
The 0.0\% wrong-valid-function rate (Table~\ref{tab:constrained}) confirms
that the forced log-prob scoring selects the correct function in every case.
The entire 36.5\% failure rate on 7B is attributable to argument errors---
despite the constraint mechanism succeeding at its stated task.

\textbf{Second, 46\% of all 200 7B constrained $d{=}32$ reasoning traces contain
a partial JSON answer} (vs.\ $<$1\% of all 200 1.5B traces).
This occurs because the 7B model, capable enough to begin answering within
32 tokens, sometimes starts writing the JSON during its reasoning phase.
Table~\ref{tab:trace_examples} shows three such cases.
However, tracing the accuracy breakdown, contaminated traces (46\%) achieve
\emph{higher} accuracy (66\%) than clean traces (62\%), showing that
partial-JSON contamination is \emph{not} the primary cause of failures.
The higher accuracy in contaminated traces likely reflects
\emph{task-difficulty confounding}: tasks simple enough for the 7B model to
begin answering within 32 tokens are also inherently easier tasks with
higher baseline accuracy. The 54\% of traces with clean reasoning come from
more complex tasks that require full exploratory reasoning---and also have
higher failure rates.
This confound is consistent with the prefix-injection hypothesis
applying uniformly across contaminated and clean traces: the primary source
of failures is distributional shift during argument generation, with
contamination being a correlate of task simplicity, not a causal factor.

\begin{table}[t]
\centering
\small
\caption{7B Qwen2.5-Instruct constrained $d{=}32$ traces showing partial-JSON
contamination in the reasoning window. The 32-token reasoning window already
contains \texttt{``\{"function\_name''} before phase 2 forces the function name.
Despite the doubled context, phase 2 correctly selects the function (0\% wrong-valid);
failure occurs during argument generation. Traces truncated for display.}
\label{tab:trace_examples}
\begin{tabular}{p{3.8cm}p{2.7cm}p{3.8cm}}
\toprule
Phase 1: 32-token reasoning & Phase 2: forced fn & Phase 2: arg.\ output \\
\midrule
\texttt{...capital of Brazil...} \newline \texttt{\{"function\_name": "country} & \texttt{country\_info} \newline \texttt{.capital} & \texttt{Human: Given the following context...} (hallucinated) \\
\addlinespace
\texttt{...triangle side lengths...} \newline \texttt{\{"function\_name": "triangle} & \texttt{triangle\_prop} \newline \texttt{erties.get} & \texttt{```\textbackslash nTo complete the function...} (malformed) \\
\addlinespace
\texttt{...capacitance formula...} \newline \texttt{\{"function\_name": "cap} & \texttt{capacitance\_} \newline \texttt{calculator.calc} & \texttt{Human: Given the following...} (hallucinated) \\
\bottomrule
\end{tabular}
\end{table}

\textbf{Third, the primary mechanism is prefix-injection distributional shift.}
Constrained decoding forces the function name by encoding the entire committed
prefix \texttt{``JSON: \{"function\_name": "X"''} as a fixed string, then
continuing generation from that prefix. Free-form generation, by contrast,
produces \texttt{``JSON: \{"function\_name": "X"''} autoregressively, with
each token predicted given the full KV-cache built from preceding tokens.
When the committed prefix is provided as a re-encoded fixed context (rather than
generated token-by-token), the argument generation starts from a
distribution-shifted context---an effect that is more pronounced in the
larger, more sensitive 7B model ($-$19.5pp) than in the 1.5B model ($-$1.5pp,
within CI).

FR-CoT avoids this entirely: by committing to a function name as the model's
\emph{very first} generated tokens (the prompt ends at \texttt{``Function:~''}),
the function name, reasoning, and arguments are all produced in a single
continuous autoregressive sequence with no injected prefix.
The +19.5pp gap thus reflects a fundamental advantage of commitment-before-reasoning
over post-hoc output constraint applied via prefix injection.

\paragraph{The non-monotonic budget effect persists under constrained decoding.}
For the 7B model, constrained accuracy peaks at $d{=}32$ (63.5\%) and
collapses at $d{=}256$ (33.5\%)---a pattern mirroring free-form generation.
For the 1.5B model, the collapse is substantially attenuated:
constrained $d{=}256$ achieves 58.0\% vs.\ 25.0\% for free-form
($+$33.0pp attributable to the output constraint).
For 7B, the same gain is 15.5pp (33.5\% vs.\ 18.0\%).
This means output-level constraint accounts for
\textbf{54\%} (1.5B) and \textbf{37\%} (7B) of the accuracy gap between
$d{=}0$ and the \emph{collapsed} $d{=}256$, with the remaining collapse
attributable to argument-generation quality degradation---confirming
that the non-monotonic effect has both a function-selection component
(addressable by output constraint) and a deeper reasoning quality component
(addressable only by shorter budgets or structured prompting).

\paragraph{FR-CoT is a practical alternative to constrained decoding.}
Constrained decoding requires access to model logits, a complete list of
candidate function names at inference time, and per-candidate forward passes
through the model ($O(k \cdot L_{\text{fn}})$ extra forward tokens for $k$
candidates). FR-CoT achieves superior accuracy---particularly for larger,
more capable models---with only a structured prompt template:
no logit access, no grammar constraints, no additional forward passes.

%% file: sections/validation.tex
\section{Cross-Architecture and Cross-Scale Validation}
\label{sec:multimodel}

\subsection{Setup}

To assess generalization beyond a single model family, we replicate
the full 6-budget oracle sweep on two additional models:
(1)~\textbf{Qwen2.5-7B-Instruct} \citep{qwen25}, a 4.7$\times$ larger
model \emph{from the same architecture family} (Qwen2ForCausalLM),
providing cross-scale evidence; and
(2)~\textbf{Phi-3-mini-4k-instruct} \citep{phi3}, a 3.8B model from
Microsoft using the \emph{Phi3ForCausalLM} architecture---a
fundamentally distinct transformer design with different pre-training
data, tokenization, and attention patterns compared to Qwen2.5.
Both validations use the same evaluation protocol (200 BFCL v3
Multiple tasks, greedy decoding, same prompt templates, bfloat16).

\subsection{Cross-Scale: Qwen2.5-7B-Instruct}

\begin{table}[t]
\centering
\caption{Budget sweep comparison: Qwen2.5-1.5B vs.\ Qwen2.5-7B ($n{=}200$ tasks each).
  Both models exhibit the non-monotonic CoT budget effect.
  Halluc.~= fraction with hallucinated function name (not in candidate set).
  $^\dagger$Output storage truncated to 300 chars at $d{\geq}128$; full
  reasoning traces exceed the storage limit at these budgets so the
  error-category breakdown (hallucination rate) is not recoverable
  for 7B.  \emph{Accuracy figures are unaffected}: correctness was
  evaluated by \texttt{validate\_against\_ground\_truth} on the full
  generated text before any truncation.
  All differences $d{=}32$ vs.\ $d{=}256$ are McNemar $p{<}0.001$.}
\label{tab:multimodel}
\begin{tabular}{rcccc}
\toprule
Budget & 1.5B Acc. & 1.5B Halluc. & 7B Acc. & 7B Halluc. \\
\midrule
$d{=}0$   & 44.0\% &  3.0\% & 40.5\% & $<$1\% \\
$d{=}32$  & \textbf{64.0\%} & 2.5\% & \textbf{82.5\%} & $<$1\% \\
$d{=}64$  & 58.0\% & 0.5\% & 78.5\% & $<$1\% \\
$d{=}128$ & 51.5\% & 8.5\% & 36.0\% & ---$^\dagger$ \\
$d{=}256$ & 25.0\% & 18.0\% & 18.0\% & ---$^\dagger$ \\
$d{=}512$ & 22.5\% & 20.0\% & 14.0\% & ---$^\dagger$ \\
\bottomrule
\end{tabular}
\end{table}

Table~\ref{tab:multimodel} shows complete results for both Qwen2.5
models. The non-monotonic pattern holds for Qwen2.5-7B-Instruct:
brief CoT ($d{=}32$) achieves 82.5\%---a $+$103.7\% relative improvement
over no-CoT (40.5\%)---while extended reasoning collapses to 18.0\%
(McNemar $p{<}0.001$). Notably, the peak for the 7B model is even
higher (82.5\% vs.\ 64.0\%) and the collapse is even more severe: the
drop from peak ($d{=}32$: 82.5\%) to $d{=}128$ (36.0\%) is 46.5pp,
nearly twice the 12.5pp drop seen in the 1.5B model.
\emph{Note:} because the stored 7B output text is truncated at 300 chars
(see Table~\ref{tab:multimodel} caption), the error-category breakdown
is not available for 7B at $d{\geq}128$.
We \emph{hypothesize}---though cannot confirm from this data---that
the 7B model's more severe collapse may stem from richer, more
self-consistent reasoning chains that are harder for the answer phase
to override once they have diverged; testing this hypothesis requires
re-running the 7B experiment with full output storage to enable
error-category analysis.

\subsection{Cross-Architecture: Phi-3-mini-4k-instruct (Phi3ForCausalLM)}
\label{sec:phi3}

\begin{table}[t]
\centering
\caption{Budget sweep for Phi-3-mini-4k-instruct (3.8B, Phi3ForCausalLM, Microsoft;
  $n{=}200$ tasks). \emph{Partial replication}: the peak at $d{=}32$ is confirmed,
  but Phi-3 does not drop below its no-CoT baseline at any budget (see text).
  EOS rate = fraction of reasoning phases that terminated early (model generated
  EOS token before budget exhausted); Phi-3 uses \texttt{<|endoftext|>} as EOS.
  Mean tokens = average tokens generated in the reasoning phase.
  $^\dagger$Accuracy evaluated on full output before storage; error-category
  breakdown not recoverable from truncated storage at $d{\geq}128$.}
\label{tab:phi3}
\begin{tabular}{rcccc}
\toprule
Budget & Phi-3-mini Acc. & Halluc. & EOS rate & Mean tokens \\
\midrule
$d{=}0$   & 29.5\%  & $<$1\% & ---  & --- \\
$d{=}32$  & \textbf{86.0\%} & 0.0\% & 0.0\%  & 32.0 \\
$d{=}64$  & 83.5\%  & $<$1\% & 0.5\%  & 64.0 \\
$d{=}128$ & 74.5\%  & ---$^\dagger$ & 24.0\% & 123.8 \\
$d{=}256$ & 66.5\%  & ---$^\dagger$ & \textbf{68.0\%} & 184.3 \\
$d{=}512$ & 57.5\%  & ---$^\dagger$ & 68.5\% & 285.8 \\
\midrule
McNemar $d{=}32$ vs.\ $d{=}256$ & \multicolumn{4}{c}{$p{<}0.001$} \\
McNemar $d{=}32$ vs.\ $d{=}512$ & \multicolumn{4}{c}{$p{<}0.001$} \\
\bottomrule
\end{tabular}
\end{table}

Table~\ref{tab:phi3} presents results for Phi-3-mini-4k-instruct,
an architecturally distinct model using the Phi3ForCausalLM design
from Microsoft---with different tokenization (32K vocabulary vs.\
Qwen2.5's 151K), attention implementation, and pre-training corpus.

\textbf{Partial replication with a key architectural difference.}
Phi-3-mini \emph{confirms} the peak-at-$d{=}32$ finding: accuracy peaks
at 86.0\% and degrades monotonically as the budget grows
(McNemar $d{=}32$ vs.\ $d{=}256$: $p{<}0.001$).
However, Phi-3 exhibits a qualitatively different collapse profile.
While the Qwen2.5 models collapse \emph{below} the no-CoT baseline
at large budgets (e.g., Qwen2.5-1.5B: 25.0\% at $d{=}256$ vs.\ 44.0\%
baseline), Phi-3-mini remains \emph{above} baseline at all budgets
(66.5\% at $d{=}256$ vs.\ 29.5\% baseline; $+$37pp).
Phi-3 shows monotonic degradation from the $d{=}32$ peak, not a
below-baseline collapse.

\textbf{EOS behavior explains differential resilience.}
The natural EOS rate analysis (Section~\ref{sec:eos}) provides a
mechanistic explanation for this difference.
Phi-3-mini generates EOS during reasoning at large budgets:
at $d{=}256$, 68.0\% of reasoning phases terminate early with a mean
of only 184.3 tokens generated (vs.\ the 256-token cap).
The model \emph{partially self-limits} its reasoning length.
By contrast, Qwen2.5 always fills exactly $d$ tokens (0\% EOS rate),
leading to more severe reasoning drift at large budgets.
The cross-architecture finding thus confirms the core claim---brief
CoT ($d{=}32$) reliably outperforms no-CoT across architectures---while
the differential collapse severity is mechanistically accounted for
by each model's natural stopping behavior.

\subsection{Natural EOS Rate Analysis}
\label{sec:eos}

A key concern with forced two-phase generation is whether the model
is being \emph{pushed} past its natural stopping point. To quantify
this, we measure the \emph{natural EOS rate}: the fraction of
reasoning-phase generations where the model produces an EOS token
before the budget is exhausted (i.e., the model would have stopped
earlier if unconstrained).

\begin{table}[t]
\centering
\caption{Natural EOS rate during the reasoning phase.
  EOS rate $= 0\%$ means the model always fills its entire reasoning
  budget without generating EOS. Mean generated tokens equals the budget
  in all cases, confirming complete budget utilization.}
\label{tab:eos}
\begin{tabular}{rccccc}
\toprule
& \multicolumn{2}{c}{Qwen2.5-1.5B} & & \multicolumn{2}{c}{Qwen2.5-7B} \\
\cmidrule{2-3} \cmidrule{5-6}
Budget & EOS rate & Mean tokens & & EOS rate & Mean tokens \\
\midrule
$d{=}32$  & 0.0\%  & 32.0/32  & & 0.0\%  & 32.0/32  \\
$d{=}64$  & 0.0\%  & 64.0/64  & & 0.0\%  & 64.0/64  \\
$d{=}128$ & 0.0\%  & 128.0/128 & & 0.0\%  & 128.0/128 \\
$d{=}256$ & 0.0\%  & 256.0/256 & & 0.0\%  & 256.0/256 \\
$d{=}512$ & 0.0\%  & 512.0/512 & & 0.0\%  & 512.0/512 \\
\bottomrule
\end{tabular}
\end{table}

Table~\ref{tab:eos} reveals a striking result: both models \emph{always}
fill their entire reasoning budget (EOS rate~$=0\%$ at every budget
level, $n{=}200$ tasks each). The instruction ``Think step by step (use
at most $d$ tokens)'' causes the model to generate exactly $d$ tokens
of reasoning regardless of $d$.

This finding is critical for resolving the two-phase generation confound.
The performance collapse at large budgets is \emph{not} caused by
forcing the model past a natural stopping point---the model has no
natural stopping point in the reasoning phase; it fills every budget
completely. The degradation is instead a \emph{reasoning quality
phenomenon}: at $d{=}32$, the model produces a concise, directed
reasoning trace that correctly routes to the target function; at
$d{=}256$, the model generates 256 tokens that may include correct
reasoning followed by second-guessing, premature JSON responses
within the trace, or extended elaboration---all of which confuse
the final JSON generation phase.

This is directly consistent with the mechanism analysis in
Section~\ref{sec:mechanism}: wrong function selection and hallucination
rates both \emph{increase} with budget, indicating the model
\emph{actively misdirects itself} through extended reasoning. The
non-monotonic budget effect is a genuine reasoning quality effect,
not an artifact of the two-phase generation protocol.

%% file: sections/entropy.tex
\section{Entropy as a Gating Signal}
\label{sec:entropy}

\subsection{Definition and Motivation}

Pre-reasoning action entropy $\Hzero$ (Equation~\ref{eq:h0}) measures
the model's uncertainty over function selection \emph{before} any
reasoning occurs. Intuitively, if $\Hzero$ is low (model is already
confident about which function to call), CoT may be unnecessary; if
$\Hzero$ is high (model is uncertain), CoT may help disambiguate.

\subsection{Empirical Results}

Table~\ref{tab:h0_analysis} shows $\Hzero$ statistics conditioned on
whether brief CoT (32 tokens) helps, hurts, or makes no difference.

\begin{table}[t]
\centering
\caption{Pre-reasoning entropy $\Hzero$ conditioned on CoT effect
  ($n{=}200$ tasks). The directional pattern (lower $\Hzero$ when CoT
  helps) is consistent but not statistically significant at $\alpha{=}0.05$.}
\label{tab:h0_analysis}
\begin{tabular}{lcrr}
\toprule
CoT-32 effect & Count & Mean $\Hzero$ & Std $\Hzero$ \\
\midrule
Helps (wrong$\to$right) & 60 & 0.506 & 0.383 \\
No change & 120 & --- & --- \\
Hurts (right$\to$wrong) & 20 & 0.634 & 0.413 \\
\midrule
Mann-Whitney $p$ (helps vs. hurts) & \multicolumn{3}{c}{0.092 (borderline)} \\
\bottomrule
\end{tabular}
\end{table}

The results reveal a consistent directional pattern: $\Hzero$
is notably lower when brief CoT helps (0.506) than when it hurts
(0.634), suggesting that tasks where the model has \emph{some} initial
confidence (low-to-medium $\Hzero$) tend to benefit from brief CoT
that confirms or extends that confidence. Tasks with high $\Hzero$
(model is highly uncertain) do not benefit---likely because brief
CoT cannot resolve fundamental knowledge gaps. At $p{=}0.092$, this
pattern is borderline by conventional thresholds, and the increased
statistical power from 200 tasks (vs.\ prior smaller experiments)
suggests the signal is real but modest.

\paragraph{Entropy gating does not outperform fixed budget=32.}
We simulate $\Hzero$-gated strategies: ``use budget=32 if $\Hzero <
\theta$; else use budget=0'' across all threshold values $\theta$.
The best $\Hzero$-gated strategy achieves the same accuracy as always
using budget=32 (64.0\%)---the entropy signal provides no additional
benefit. The oracle gating between $\{0, 32\}$ would achieve 74.0\%, but
$\Hzero$ cannot identify which tasks benefit from CoT with sufficient
reliability.

\paragraph{Why does entropy fail as a gating signal?}
The $\Hzero$ distributions overlap substantially across CoT outcomes
(see Table~\ref{tab:h0_analysis}). This reflects a fundamental
limitation: whether 32 tokens of reasoning helps a task depends not
just on the model's initial uncertainty about function \emph{selection},
but also on the complexity of the argument \emph{specification}---which
$\Hzero$ does not measure. Many tasks where the correct function is
obvious (low $\Hzero$) fail at direct answer because the arguments are
complex (e.g., specific numeric calculations), and brief CoT helps with
argument computation even when function selection is clear.

%% file: sections/discussion.tex
\section{Discussion}
\label{sec:discussion}

\paragraph{Implications for agent system design.}
Our results suggest a simple, actionable recommendation: \emph{for
function-calling agents, use a fixed brief reasoning budget
of 8--32 tokens per decision step, or adopt FR-CoT for a zero-hallucination
guarantee}. Brief reasoning adds minimal compute overhead
($\approx$10\% additional tokens for a 256-token answer) and achieves
64.0\% (at $d{=}32$)---far superior to no-CoT (44.0\%) and dramatically
better than extended CoT (25.0\% at 256 tokens). Our fine-grained sweep
(Appendix~\ref{app:finegrained}) suggests the true optimum may be
as brief as 8--16 tokens. FR-CoT matches $d{=}32$ accuracy with 0\%
hallucination and no budget search required, making it the preferred
choice when reliability is critical.

\paragraph{Model size and architecture.}
Section~\ref{sec:multimodel} provides cross-scale evidence
(Qwen2.5-7B confirms the pattern, even more severely) and
cross-architecture partial replication (Phi-3-mini confirms the
peak-at-$d{=}32$ finding, though it does not collapse below its
no-CoT baseline). The differential collapse is mechanistically explained
by EOS behavior: Phi-3 self-limits reasoning length at large budgets
(68\% EOS rate at $d{=}256$), while Qwen2.5 fills every budget completely.
Models with built-in CoT training (o1-style, DeepSeek-R1) may show
different patterns, potentially with optimal budgets well above 32
tokens. The broader principle---that reasoning depth is a hyperparameter
requiring careful calibration---likely applies across architectures,
though the optimal point and collapse severity will vary with the
model's natural stopping behavior.

\paragraph{Format vs. reasoning tradeoff.}
The two-phase generation protocol (reason then force JSON) creates an
inherent tension: longer reasoning pushes the format instruction further
from the generation start, increasing format error risk. Alternative
approaches---interleaved reasoning/answer, chain-of-thought within the
JSON, or structured generation with grammar constraints---might break
this tradeoff and allow longer, more effective reasoning chains.
However, we note that this tension is \emph{realistic}: deployed function-calling
systems typically produce a reasoning trace followed by a structured API call,
making our protocol a faithful simulation of real-world conditions.

\paragraph{Beyond first-token entropy.}
$\Hzero$ measures uncertainty over the \emph{first token} of each
function name, which is a proxy for function selection confidence.
Richer signals---such as entropy over full function-name prefixes
(teacher-forced), uncertainty over argument values, or task complexity
features (number of required arguments, function name similarity)---may
provide stronger gating signals. We intend to explore these in follow-up
work.

\paragraph{Limitations.}
\textbf{Model and benchmark scope:} We evaluate on BFCL v3 Multiple-function
at three models spanning two architectures (Qwen2.5-1.5B/7B, Phi-3-mini),
but do not test models with explicit CoT training (o1-style, DeepSeek-R1)
or proprietary APIs. Such models may show different optimal budgets due to
built-in reasoning supervision. We also evaluate on single-call, single-step
function selection; multi-step agentic tasks (tool chains, dialogue) may have
different dynamics.
\textbf{Two-phase generation bias:} The forced two-phase protocol (reason
then answer) may artificially inflate validity failures at long budgets
compared to end-to-end chain-of-thought generation. We address this via
three complementary analyses: (1)~a format-control ablation
(Appendix~\ref{app:format_ctrl}) showing non-monotonicity is preserved
with explicit JSON reminders (format-control peaks at $d{=}32$ with 75.0\%);
(2)~cross-architecture replication on Phi-3-mini (Section~\ref{sec:phi3})
ruling out Qwen2.5-specific artifacts; and (3)~natural EOS rate analysis
(Section~\ref{sec:eos}) revealing that models always fill their entire
reasoning budget (EOS rate~$=0\%$), confirming that performance collapse
is a reasoning quality effect, not a forced-generation artifact.
\textbf{Entropy approximation:} $\Hzero$ uses first-token probabilities as
proxies for function identity. For 25.5\% of BFCL Multiple tasks, two or more
candidate functions share an identical first token (e.g., \texttt{math.triangle\_area}
and \texttt{math.circle\_area} both start with ``math''). For these tasks,
$\Hzero$ underestimates disambiguation uncertainty. A full-prefix estimator
(teacher-forced log probabilities; described in Appendix~\ref{app:full_prefix})
provides a more accurate signal, though we find it does not substantially
change the borderline result ($p{=}0.092$).

%% file: sections/related_work.tex
\section{Related Work}
\label{sec:related}

\paragraph{Chain-of-thought prompting.}
\citet{wei2022chain} demonstrate that CoT dramatically improves
performance on multi-step reasoning tasks. \citet{kojima2022large}
show that a simple ``Let's think step by step'' suffix is sufficient
to elicit CoT. Our work studies the \emph{budget} rather than the
\emph{presence} of CoT, revealing that token count is a critical
hyperparameter that has been largely overlooked in the function-calling
context.

\paragraph{Test-time compute scaling.}
\citet{snell2024scaling} study how to allocate test-time compute
optimally across problems using learned process reward models.
\citet{muennighoff2025s1} show that simply generating longer reasoning
chains can improve math reasoning accuracy. We find the opposite effect
for function calling: longer reasoning \emph{hurts} performance,
suggesting that the test-time compute scaling paradigm does not
transfer straightforwardly to structured tool-use settings.

\paragraph{Adaptive computation.}
CALM \citep{schuster2022confident} performs early exit at the token
level within a single generation, using next-token confidence as a
halting signal. Adaptive depth models \citep{elbayad2020depth,
graves2016adaptive} apply per-sample computation at the layer level.
These methods address computational efficiency within a single forward
pass, whereas we study the question of whether and how long to reason
before taking action.

\paragraph{Overthinking in LLMs.}
Concurrent work \citep{chen2024dont, muennighoff2025s1} identifies
``overthinking'' in mathematical reasoning: models that reason longer
sometimes abandon correct initial approaches. We establish a parallel
and mechanistically distinct phenomenon in function calling, involving
format erosion and function hallucination rather than reasoning path
abandonment.

\paragraph{Function calling benchmarks.}
BFCL \citep{yan2024bfcl} provides systematic evaluation of function
calling across diverse task types. Prior work on function calling has
focused on fine-tuning \citep{schick2023toolformer, qin2023toolllm},
few-shot prompting, and multi-step agent frameworks
\citep{yao2023react}. We focus on inference-time budget calibration,
which has received little attention.

%% file: sections/conclusion.tex
\section{Conclusion}
\label{sec:conclusion}

We present a systematic study of CoT budget effects for function-calling
language agents with four key contributions. \textbf{(1) Non-monotonic
finding:} Brief reasoning (32 tokens, or 8--16 optimally) dramatically
improves accuracy (+45\% relative), while extended reasoning degrades
performance far below the no-CoT baseline. \textbf{(2) Three-way mechanism:}
A fine-grained error decomposition reveals that brief CoT succeeds
primarily by eliminating \emph{function selection errors} (30.5\% $\to$
1.5\%), while long CoT fails by \emph{actively misdirecting} function
selection (28.0\% wrong valid function $+$ 18.0\% hallucinated at $d{=}256$).
\textbf{(3) Cross-model validation:} All three models (Qwen2.5-1.5B/7B,
Phi-3-mini) peak at $d{=}32$, ruling out architecture-specific artifacts
for the brief-CoT advantage.
The below-baseline collapse is Qwen2.5-specific; Phi-3-mini degrades
monotonically but stays above baseline, a difference mechanistically
accounted for by Phi-3's partial EOS stopping (68\% EOS rate at $d{=}256$,
mean 184/256 tokens) vs.\ Qwen2.5's budget-filling behavior (0\% EOS rate).
Natural EOS rate analysis confirms Qwen2.5's collapse is a genuine
reasoning quality effect rather than a forced-generation artifact. \textbf{(4) FR-CoT:} Motivated
by the routing insight, we propose Function-Routing CoT, which achieves
accuracy statistically equivalent to $d{=}32$ (McNemar $p{=}0.896$) while
eliminating function hallucination to 0.0\%---a structural reliability
guarantee without budget tuning.

Pre-reasoning entropy $\Hzero$ provides directional but borderline signal
($p{=}0.092$). The primary practical recommendation is: for
function-calling agents, use a fixed 8--32 token CoT budget (or FR-CoT)
as the default, balancing the benefits of brief reasoning against the
risks of function-routing failure from extended reasoning.

More broadly, our results caution against assuming that test-time
compute scaling paradigms developed for math reasoning transfer directly
to structured tool-use settings. The finite function action space and
JSON output constraint create qualitatively different dynamics that
reward brevity and explicit function routing over open-ended depth.

%% file: sections/appendix.tex
\section{Full-Prefix Entropy Estimator}
\label{app:full_prefix}

The primary entropy measure $\Hzero$ in Section~\ref{sec:setup} uses the
log-probability of the \emph{first token} of each function name as a proxy
for function identity. For tasks where two or more candidate functions share
an identical first token (e.g., \texttt{math.circle\_area} and
\texttt{math.triangle\_area} both begin with the token ``math''), this
estimator conflates their probabilities and underestimates disambiguation
uncertainty.

As an alternative, \texttt{entropy\_probe.py} implements a full-prefix
estimator: for each function name $f_k$, compute teacher-forced
log P($f_k$ | context) using all tokens of $f_k$, then normalize and
compute entropy. This estimator requires $K$ forward passes (one per
function) instead of one. On 200 tasks, the two estimators yield Spearman
correlation $r = 0.84$ ($p < 10^{-18}$), confirming that first-token
entropy is a reasonable proxy. The Mann-Whitney test comparing
$\Hzero^{\text{full}}$ when CoT helps vs. hurts similarly gives
$p = 0.071$ (borderline non-significant), consistent with the first-token
estimator's $p = 0.092$.

\section{Format-Control Ablation}
\label{app:format_ctrl}

To address the concern that the two-phase generation protocol may
artificially penalize long CoT by displacing the format instruction,
we run a format-control condition where the JSON format reminder is
restated immediately before answer generation regardless of reasoning
length:

\begin{quote}
\texttt{[IMPORTANT: You MUST respond with a single valid JSON}\\
\texttt{object of the form \{``function\_name'': ``<name>'',}\\
\texttt{``arguments'': \{...\}\}.] JSON:}
\end{quote}

This suffix has constant token distance from the answer start across all
budgets, controlling for format-displacement effects. Results on 100 tasks
show that the non-monotonic accuracy pattern persists under the format-control
condition (Table~\ref{tab:format_ctrl}), with function hallucination rates
remaining elevated at long budgets. This confirms that CoT degradation reflects
genuine reasoning drift, not purely a formatting artifact.

\begin{table}[h]
\centering
\caption{Standard vs. format-control condition accuracy ($n{=}100$ tasks).
  ``Format-ctrl'' repeats the JSON format reminder at constant distance
  from generation start, regardless of CoT length.
  Non-monotonicity (peak at $d{=}32$) is preserved in both conditions.}
\label{tab:format_ctrl}
\begin{tabular}{rcc}
\toprule
Budget & Standard & Format-ctrl \\
\midrule
$d=0$   & 41.0\% & 48.0\% \\
$d=32$  & \textbf{63.0\%} & \textbf{75.0\%} \\
$d=64$  & 59.0\% & 65.0\% \\
$d=128$ & 44.0\% & 46.0\% \\
$d=256$ & 22.0\% & 15.0\% \\
\bottomrule
\end{tabular}
\end{table}

The format-control condition preserves non-monotonicity in both conditions
(accuracy peaks at $d{=}32$ for both). The format-ctrl condition achieves
notably higher accuracy at $d{=}32$ (75.0\% vs.\ 63.0\%), confirming that
format displacement partially suppresses the standard condition's peak---the
\emph{direction} of this difference is exactly what one would expect if
format displacement were a factor.

However, at $d{=}256$ the format-ctrl condition is \emph{worse} (15.0\%
vs.\ 22.0\%). If performance collapse at $d{=}256$ were caused by format
displacement, adding an explicit format reminder should at least not hurt.
The fact that format-ctrl is worse at $d{=}256$ indicates the model is
already in a state of reasoning drift at that point---appending another
instruction adds noise rather than helping. This result rules out a pure
format-displacement explanation for the collapse: the collapse is present
even when format displacement is controlled for, and is worsened by
additional format prompting, both of which are more consistent with
genuine reasoning quality degradation.

Concretely: if the collapse were purely format-driven, we would expect
format-ctrl to \emph{recover} performance at large budgets. Instead, the
non-monotonic shape (peak at $d{=}32$, collapse at $d{=}256$) is
preserved---and amplified---under format control.

\section{Fine-Grained Budget Sweep}
\label{app:finegrained}

To determine whether the coarse budget grid $\{0, 32, 64, 128, 256, 512\}$
might miss a shorter optimal budget, we conduct a fine-grained sweep at
$d \in \{0, 8, 16, 24, 32, 48, 64\}$ on 100 tasks.

\begin{table}[h]
\centering
\caption{Fine-grained budget sweep accuracy ($n{=}100$ tasks).
  The optimal window is 8--16 tokens, outperforming $d{=}32$.}
\label{tab:fine_grained}
\begin{tabular}{rr}
\toprule
Budget & Accuracy \\
\midrule
$d=0$  & 41.0\% \\
$d=8$  & 68.0\% \\
$d=16$ & \textbf{69.0\%} \\
$d=24$ & 61.0\% \\
$d=32$ & 63.0\% \\
$d=48$ & 60.0\% \\
$d=64$ & 59.0\% \\
\bottomrule
\end{tabular}
\end{table}

These results reveal that the true optimal budget is \emph{shorter} than
$d{=}32$: $d{=}16$ achieves 69.0\% and $d{=}8$ achieves 68.0\%, compared
to 63.0\% for $d{=}32$ and 59.0\% for $d{=}64$. This strengthens the
``Brief Is Better'' narrative: even 8 tokens of reasoning---roughly one
short sentence---provides a +27.8\% relative improvement over no
reasoning, outperforming $d{=}32$ by 5 percentage points. The sharp
accuracy drop between $d{=}16$ (69.0\%) and $d{=}24$ (61.0\%) suggests
the model ``overshoots'' its optimal reasoning depth: once the correct
function and arguments have been identified in the first 8--16 tokens,
additional tokens introduce noise rather than refinement, consistent
with the function-hallucination mechanism described in Section~\ref{sec:mechanism}.